\documentclass[conference]{IEEEtran}
\IEEEoverridecommandlockouts
\usepackage{cite}
\usepackage{balance}
\usepackage{amsmath,amssymb,amsfonts}
\usepackage{algorithmic}
\usepackage{graphicx}
\usepackage{textcomp}
\usepackage{xcolor}
\usepackage{booktabs, multicol, multirow}
\usepackage{array}
\def\BibTeX{{\rm B\kern-.05em{\sc i\kern-.025em b}\kern-.08em
    T\kern-.1667em\lower.7ex\hbox{E}\kern-.125emX}}
\begin{document}

%\title{Improving Multibiometric Authentication System Performance using Neural Network Decoder \\

\title{Learning to Authenticate with Deep Multibiometric Hashing and Neural Network Decoding \\
}

\author{\IEEEauthorblockN{Veeru Talreja, Sobhan Soleymani, Matthew C. Valenti, and Nasser M. Nasrabadi  }
\IEEEauthorblockA{Lane Department of Computer Science and Electrical Engineering \\
West Virginia University, Morgantown, WV, USA\\
Email: vtalreja@mix.wvu.edu, ssoleyma@mix.wvu.edu, valenti@ieee.org, and nasser.nasrabadi@mail.wvu.edu}
}

\maketitle

\begin{abstract}
%Multimodal authentication systems have numerous advantages when compared to unimodal systems such as lower error rate, higher accuracy, and larger population coverage. 
In this paper, we propose a novel multimodal deep hashing neural decoder (MDHND) architecture, which integrates a deep hashing framework with a neural network decoder (NND) to create an effective multibiometric authentication system. The MDHND consists of two separate modules: a multimodal deep hashing (MDH) module, which is used for feature-level fusion and binarization of multiple biometrics, and a neural network decoder (NND) module, which is used to refine the intermediate binary codes generated by the MDH and compensate for the difference between enrollment and probe biometrics (variations in pose, illumination, etc.). Use of NND helps to improve the performance of the overall multimodal authentication system. The MDHND framework is trained in 3 steps using joint optimization of the two modules. In Step 1, the MDH parameters are trained and learned to generate a shared multimodal latent code; in Step 2, the latent codes from Step 1 are passed through a conventional error-correcting code (ECC) decoder to generate the ground truth to train a neural network decoder (NND); in Step 3, the NND decoder is trained using the ground truth from Step 2 and the MDH and NND are jointly optimized. Experimental results on a standard multimodal dataset demonstrate the superiority of our method relative to other current multimodal authentication systems. 

%Furthermore,  the  proposed system can  work  in  both  identification  and  authentication  modes.      

\end{abstract}

\begin{IEEEkeywords}
deep hashing, multibiometric, error-correcting codes, authentication
\end{IEEEkeywords}

\section{Introduction}
Multimodal biometric authentication systems use a combination of different biometric traits such as face and iris, or face and fingerprint to authenticate the user. Multimodal systems have the advantage of lower error rates and higher accuracy when compared to unimodal systems \cite{nagar_multibiometriccryptosystems_2012}. Additionally, multimodal systems are generally more resistant to spoofing attacks, and can be made to be more universal than unimodal systems, since the use of multiple modalities can compensate for missing modalities in a small portion of the population \cite{ross_multibiometric_2004}. 

%Consequently, multimodal systems have been deployed in many large scale biometric applications including the FBI's Next Genration Identification (NGI), the Department of Homeland Security's US-VISIT, and the Government of India's UID.
%Sensor-level, score-level and decision-level fusion have been extensively studied in the literature \cite{Aguilar_2003_fusion,Hong_1998_integrating}.
One of the major challenges of multimodal systems is the selection of the fusion algorithm required to combine the multiple modalities. The fusion of the modalities can be performed  at sensor, feature, score, or  decision levels \cite{ross_multibiometric_2004,nandakumar_multibiometric_2008,taherkhani2018deep,talreja_multibiometric}. However, fusion-level authentication systems have better performance because they leverage the richer information available about the biometric data in the feature-set \cite{Xin_2018_multimodal}. Feature-level fusion involves combining of feature vectors that are
obtained from multiple feature sources. Multibiometric systems can combine biometrics in many ways, including: (i) feature vectors obtained from different sensors for the same biometric; (ii) feature vectors obtained from different entities for the same biometric, such as combination of feature vectors obtained from left and right eyes; or (iii) feature vectors obtained from multiple biometric traits, such as face and iris. Fusion at the feature level is relatively difficult to achieve in practice due to incompatibility of feature sets extracted from multiple modalities and because the relationship between different feature spaces may be unknown \cite{shi_2016_rule}. To overcome this issue, we propose a feature-level fusion of multiple biometrics using convolutional neural networks (CNNs). 

%Multibiometric systems can combine biometrics in many ways, including: (i) feature vectors obtained from different sensors for the same biometric; (ii) feature vectors obtained from different entities for the same biometric, such as combination of feature vectors obtained from left and right eyes; and (iii) feature vectors obtained from multiple biometric traits, such as face and iris. 

%Moreover, simple concatenation of feature sets may lead to the problem of curse of dimensionality and a very complex matcher may be required to operate on the concatenated feature set \cite{ross_feature_2005,rattani_2007_feature}. 
%because the relationship between different feature spaces may be unknown. Moreover, as the feature space dimensionality increases, the dependency between the modalities is not exploited efficiently \cite{shi_2016_rule}

%Recent progress in image classification, object detection, face recognition, speech recognition, and many other computer vision tasks demonstrates the impressive learning ability of CNNs. CNNs have also been utilized as feature extractors for biometric modalities such as face, iris, and fingerprint \cite{nakada_2017_acfr,Gangwar_2016_Deepiris,lin_2019_cnn}. The robustness of features generated by CNNs has led to a surge in the application of CNN-based deep learning methods for generating binary (hash) codes from raw image data. Deep hashing is the application of deep learning to generate compact binary vectors from raw image data \cite{supervised_xia_2014,learning_lin_2016,deep_hashing_liu_2016}. 

In this paper, we have combined multiple modalities at feature-level using two different CNN architectures. Similar work has been done in \cite{sobhan_2018_icpr}, however, the CNN features are high-dimensional and real-valued, and usually require high computational cost \cite{Taherkhani_IET_2018}. In order to reduce the computational complexity, in this paper, we have also integrated a novel deep hashing algorithm into a feature-level fusion CNN architecture to design a multimodal deep hashing (MDH) network for combining multiple modalities. Deep hashing is the application of deep learning to generate compact binary vectors from raw image data \cite{talreja_globalsip_2018,taherkhani_2018_biosig}. In addition to using deep hashing for feature-level fusion, we have also used error correcting codes (ECC) as an additional component to compensate for the difference in enrollment and probe biometrics (arising from variation in pose, illumination, noise in biometric capture). Due to this difference, enrollment and probe biometrics lead to different hash code at the output of the MDH network and therefore a failure to authenticate.  ECC can compensate for this difference by forcing the enrollment and probe biometric to decode to the same message, then using that message for authentication, making the system more robust to distortion in the biometric measurements, which helps in improving the authentication performance. 

Recent work has shown that the same kinds of neural network architectures used for classification can also be used to decode ECC codes \cite{2016_nachmani_NND}. In this work, we integrate a neural network decoder (NND) \cite{2016_nachmani_NND} into our deep hashing architecture as an ECC component to improve the authentication performance. The NND is a formulation of the belief propagation (BP) algorithm as a neural network. The input to the NND can be considered to be a corrupted codeword of an  ECC and this corrupted codeword is within a certain distance of a correct codeword of an ECC. The corrupted codeword can be decoded using the NND to generate the correct codeword. It can be argued that a conventional ECC decoder can be used instead of using a NND, however, implementing the decoder as a neural network has the benefit of providing the same architecture as the classifier (MDH network) that extracts the hashing code, and hence, it can be more efficiently jointly optimized and implemented within a common framework. Another advantage with NND is that it provides an opportunity to jointly learn and optimize with respect to biometric datasets, which are not necessarily characterized simply by Gaussian noise as is assumed by a conventional decoder. Motivated by this, in this paper, we have integrated our MDH network with NND and a joint optimization process to formulate our novel MDHND framework for an end-to-end multimodal biometric authentication system. 

To summarize the main contributions of this paper include:
\begin{enumerate}
    \item Conversion of the involved biometric traits into a common feature space by using modality-specific CNNs. 
    \item Fusion and binarization of the individual modality features to generate a robust binary latent shared representation.
    \item Inclusion and optimization of the neural network based decoder to compensate for the distortion in biometric measurements.
    \item End-to-end joint optimization of the overall system.
    %\item Application of the system in the domains of authentication as well as identification, where the size of the supported identity database depends on the strength of the ECC. 

\end{enumerate}

\section{Multimodal Deep Hashing Neural Decoder}\label{sec:mdhnd}

%\subsection{System Overview}

\begin{figure}[b]
\vspace{-0.30cm}
\centering
\includegraphics[width=8.8cm]{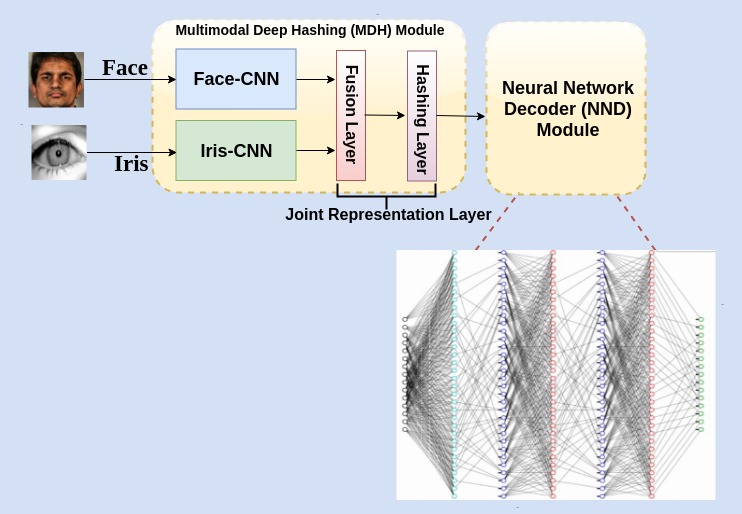}
%\vspace{-0.75cm}
\caption{Block diagram of the proposed system. The NND is shown in the figure. The MDH module is shown in Fig. \ref{fig:arch_bla}}\label{fig:enrol}
\vspace{-0.45cm}

\end{figure}

In this section, we present a system overview of the multimodal deep hashing neural decoder (MDHND) shown in Fig. \ref{fig:enrol}. MDHND consists of two important modules: multimodal deep hashing (MDH) module and neural network decoder (NND) module. We have considered face and iris as the two modalities for this authentication system. However this system could be used with other modalities and can be extended to more than two modalities.

\subsection{Multimodal Deep Hashing Module} \label{subsec:mdh}

\begin{figure}[t]
\vspace{-0.30cm}
\centering
\includegraphics[width=8.8cm]{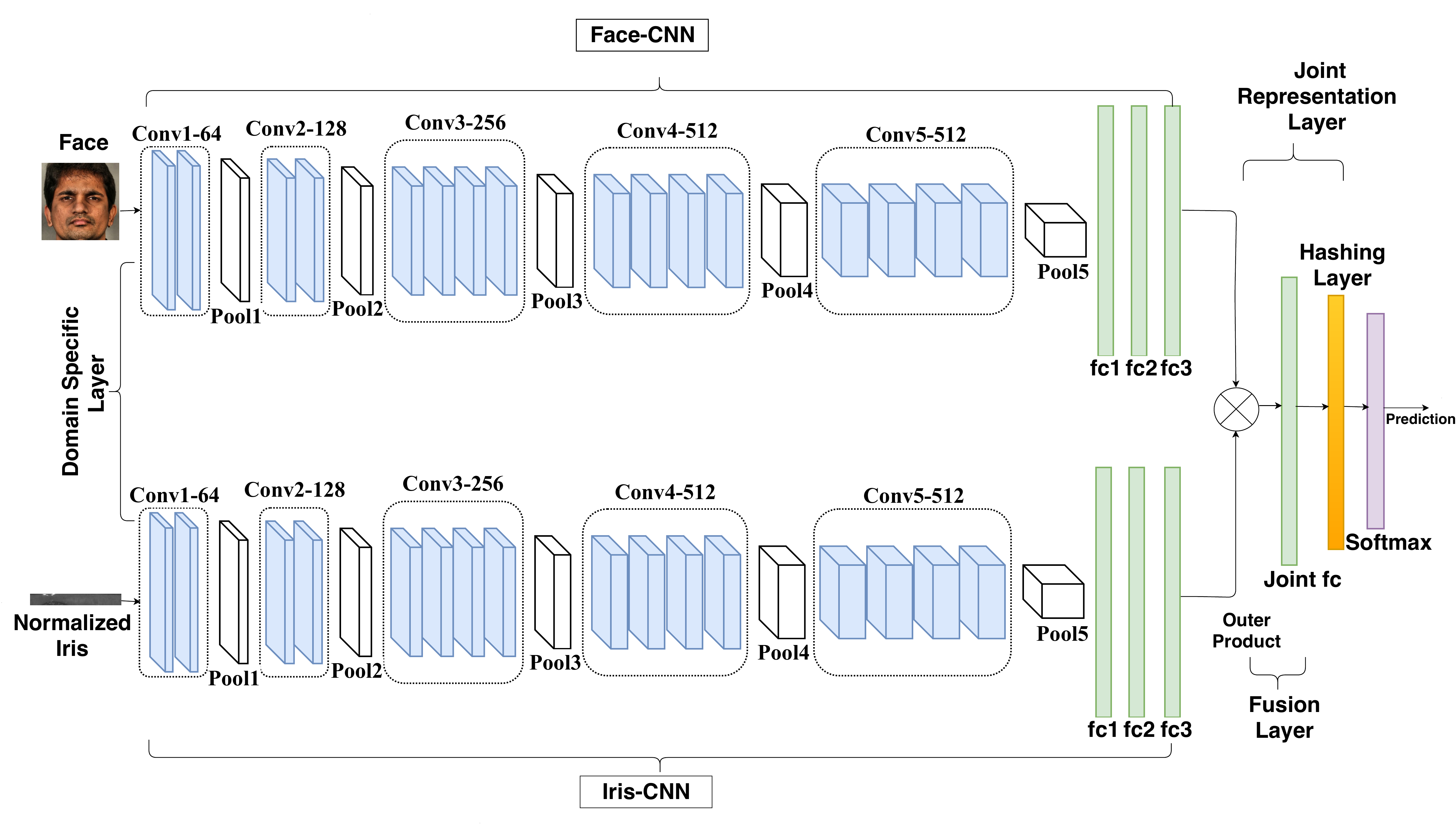}
%\vspace{-0.75cm}
\caption{Proposed multimodal deep hashing (MDH) framework for the bilinear architecture (BLA).}\label{fig:arch_bla} 
\vspace{-0.45cm}
\end{figure}

The multimodal deep hashing (MDH) module, which is shown in Fig. \ref{fig:arch_bla}, consists of a domain-specific layer containing face and iris CNNs and the joint representation layer. The primary functions of the MDH module are the non-linear feature-level fusion and  binarization of the fused features. 
%For each CNN, we use VGG-19 \cite{simonyan_very_deep_2014} pre-trained on ImageNet \cite{deng_imagenet_2009} as our starting point and then fine-tune the entire VGG-19 with an additional fully connected layer \emph{fc3}.

The domain-specific layer of the MDH module consists of a CNN for encoding the face (``Face-CNN") and another CNN for encoding the iris (``Iris-CNN"). The output feature vectors of the face and iris CNNs are fused and binarized in the joint representation layer (JRL), which is split into two sub-layers: a fusion layer and a hashing layer. The main function of the fusion layer is to fuse the individual face and iris representations from domain-specific layers into a non-linear multimodal feature embedding. The hashing layer binarizes the shared multimodal feature representation that is generated by the fusion layer.

\textbf{Fusion layer}: We have implemented two different architectures for the fusion layer: Fully Connected architecture (FCA) and Bilinear architecture (BLA). In the FCA, the outputs of the Face-CNN and Iris-CNN are concatenated vertically in the concatenation layer and passed through a fully connected layer to fuse the iris and face features. In FCA, the concatenation layer and the fully connected layer together constitute the fusion layer. In the BLA (Fig. \ref{fig:arch_bla}), the outputs of the Face-CNN and Iris-CNN are combined using the matrix outer product of the face and iris feature vectors to create bilinear feature vector. Similar to FCA, the bilinear feature vector is also passed through a fully connected layer. In BLA, the outer product layer and the fully connected layer together constitute the fusion layer.

\textbf{Hashing layer}: The output of the fusion layer is a shared multimodal feature vector of unquantized values. The output of the fusion layer can be directly binarized by thresholding at any numerical value or thresholding at the population mean. However, this kind of thresholding leads to a quantization loss, which results in sub-optimal binary codes. To account for this quantization loss, we have included another latent layer after the fusion layer, which is known as the hashing layer (shown in orange in Fig. \ref{fig:arch_bla}). The main function of the hashing layer is to capture the quantization loss incurred while converting the shared multimodal representation (output of fusion layer) into binary codes.

To generate the binary hash codes, we can directly use the sign activation function $h=\mbox{sgn}(z)$ for the hashing layer. However, the use of the non-smooth sign-activation function makes standard back-propagation impracticable as the gradient of the sign function is zero for all non-zero inputs. We have used a continuation method to overcome this zero-gradient problem by starting with a smooth activation function $y=\mbox{tanh}(\beta x)$ and making it gradually sharper by increasing the bandwidth $\beta$ as the training proceeds. This continuation utilizes the relationship between the sign activation function and the scaled $\mbox{tanh}$ function: \begin{equation}\lim_{\beta\to\infty} \mbox{tanh}(\beta x )=\mbox{sgn}(x),\label{eq:1}\end{equation} where $\beta >0$ is a scaling parameter. For training the network, we start with a $\mbox{tanh}(\beta x)$ activation for the hashing layer with $\beta=1$ and continue training until the network converges to zero loss. We then increase the value of $\beta$ while holding other training parameters equal to the previously converged network parameters, and start retraining the network. This process is repeated several times by increasing the bandwidth of the $\mbox{tanh}$ activation allowing $\beta\to\infty$ until the hashing layer can generate an output very close to binary values. In addition to using this continuation method, the overall objection function used for training the deep hashing network is discussed in Sec. \ref{subsec:stage1}

\subsection{Neural Network Decoder Module}

The output of the MDH module after training can be binarized by directly using a $\mbox{sign}$ function. Henceforth, we will refer to the output of the MDH network as \emph{intermediate binary code}. Even though we still use a threshold of $0.5$, the quantization loss is much lower and authentication performance is improved when compared to values with no hashing layer. However, there is still room to make the system more robust and improve the performance. This is achieved by using ECC. There could be distortion in biometric measurements such as variations in pose, illumination, or noise due to the biometric capturing device, which leads to difference in enrollment and probe biometrics. Due to these differences, enrollment and probe biometrics may lead to different hash codes at the output of the MDH network and therefore a failure to authenticate.  ECC can compensate for this difference by forcing the enrollment and probe biometric to decode to the same message, then using that message for authentication, making the system more robust to noise in the biometric measurements, which helps in improving the authentication performance. 

%In addition to using deep hashing for feature-level fusion, we have also used error-correcting codes (ECC) as an additional component to compensate for the noise in biometric measurements (variation in pose, illumination, etc.), .  The intermediate binary code generated by MDH network for the feature-level fusion of different modalities of the same subject can be considered to be a corrupted codeword within a certain distance of a correct codeword of an ECC. An ECC decoder can be used to reduce the distance between the binary code and the corresponding correct codeword of the ECC.  We can directly use an conventional ECC decoder. However, that would be a post-processing step and it would not be an end-to-end system. Additionally, it may not be computationally efficient to use error correcting codes with low code rate and higher error correcting capability.    

Recent research in the field of error-correcting codes (ECC) has focused on designing a neural network architecture as an ECC decoder \cite{2016_nachmani_NND,Lugosch_2017_NeuralOM}. We can adapt such a neural network based ECC decoder, train it, and use it as an ancillary component to refine the intermediate binary codes generated by MDH. The advantage of using NND instead of a conventional decoder is that it allows for a common architectural framework to be used for both the classifier (i.e., the MDH) and the decoder and it provides an opportunity to jointly learn and optimize with respect to biometric datasets, which are not necessarily characterized simply by Gaussian noise as is assumed by conventional decoders. For our application, we have chosen the NND described in \cite{2016_nachmani_NND} as the basis for our neural network ECC decoder to be integrated with the MDH network. Due to space limitations, we do not provide full details on the operation of NND, details can be found in the original paper. Rather, we focus the discussion on the changes in how the decoder is trained relative to \cite{2016_nachmani_NND}.
  
\section{Training Steps for Multimodal Deep Hashing Neural Decoder}

The MDHND framework described in Sec. \ref{sec:mdhnd} is trained in 3 steps. In Step 1, we use a novel loss function to train and learn the MDH parameters to generate binary shared multimodal latent code; in Step 2, the latent binary codes from Step 1 are passed through an ECC decoder to generate the ground truth, which will be used to fine-tune a neural network decoder (NND); in Step 3, the NND decoder is trained using the ground truth from Step 2 and this NND is then integrated with the MDH network followed by a joint optimization of the overall system MDHND. In this section, we discuss the 3 steps used during training of the proposed MDHND system.

\subsection{Step 1: Training of the Multimodal Deep Hashing Module} \label{subsec:stage1}

In this step, the MDH network is trained for feature-level fusion and binarization of the multiple biometrics. For modality-specific CNNs, VGG-19 pre-trained on ImageNet dataset is used as a starting point followed by fine-tuning the entire VGG-19 with additional fully connected layer $fc3$. The Face-CNN is fine-tuned end-to-end with the CASIA-Webface \cite{yi_learning_2014}, which contains 494,414 facial images corresponding to 10,575 subjects. The Iris-CNN is fine-tuned end-to-end using the combination of the CASIA-Iris-Thousand \cite{biometrics_ideal_test_2017} and ND-Iris-0405 \cite{bowyer_ndiris_2010} datasets with about 84,000 iris images corresponding to 1355 subjects.The Face-CNN and Iris-CNN are also fine tuned on the 2013 face and iris subsets of the WVU-Multimodal 2012-2013 datasets\cite{wvu_multimodal_2017} respectively. The WVU-Multimodal dataset for the year 2012 and 2013 together contain a total of 119,700 facial images and 257,800 iris images corresponding to 2263 subjects with 294 common subjects. For fine-tuning the Face-CNN and Iris-CNN, we have used 58,200 facial images and 121,200 iris images, respectively, corresponding to 1,060 subjects from the WVU-Multimodal 2012-2013 dataset. For fine-tuning the CNNs, we have not used any of the common subjects from 2012-2013 WVU dataset, so that these common subjects can be used for fine-tuning the NND and also for testing the overall system. 

For the Face-CNN, all the raw facial images are first aligned in 2-D and cropped to a size of $224\times 224$ before passing through the network \cite{dlib_09}. The only other pre-processing is subtracting the mean RGB value, computed on the training set, from each pixel. The training is carried out by optimizing the multinomial logistic regression objective using mini-batch gradient descent with momentum. The number of nodes in the last fully connected layer \emph{fc3} before the softmax layer is 1,024 for the FCA and 64 for the BLA. This implies that the feature vector that is extracted from the Face-CNN and fused with the feature vector from Iris-CNN has 1,024 dimensions for the FCA and 64 for the BLA. For the iris-CNN, all the raw iris images are segmented and normalized to a fixed size of $64 \times 512$ using Osiris (Open Source for IRIS) which is an open source iris recognition system developed in the framework of the BioSecure project \cite{sutra_2013_biometric}. As with the Face-CNN, the iris network has an output of 1,024 for FCA and 64 for BLA.  

Next, we train the joint representation layer, which is a combination of the fusion layer and the hashing layer. For training the joint representation layer we have used a novel objective function, which helps in reducing the quantization loss and also maximize the entropy to generate optimal and discriminative binary shared multimodal representation. For training the joint representation layer, we adopt a two-step training procedure where we first train only the joint representation layer greedily by freezing the face and iris CNNs followed by fine tuning the entire MDH module end-to-end using back-propagation at a relatively small learning rate. 

The objective function used for training the joint representation layer is a combination of classification loss, quantization loss and entropy maximization loss. The classification loss has been added into the MDH network by using the \emph{softmax} layer as shown in Fig. \ref{fig:arch_bla}. Let $E_{1}(\textbf{w})$ denote the objective function required to fulfill the classification task: 
\vspace{-0.20cm}
\begin{equation} E_{1}(\textbf{w})=\frac{1}{N}\sum_{n=1}^{N}L_n(f(x_{n},\textbf{w}),y_{n}) + \lambda ||\textbf{w}||^{2} \label{eq:2},\end{equation} where the first term $L_n(.)$ is the classification loss for a training instance $n$ and is described below, $N$ is the number of training images in a mini-batch. $f(x_{n},\textbf{w})$ is the predicted softmax output of the network and is a function of the input training image $x_n$ and the weights of the network \textbf{w}. The second term is the regularization function where $\lambda$ governs the relative importance of the regularization. 
Let the predicted softmax output $f(x_{n},\textbf{w})$ be denoted by $\hat{y}_{n}$. The classification loss for the $n^\mathsf{th}$ training instance is given as: 

\vspace{-0.20cm}
\begin{equation}L_{n}(\hat{y}_{n},y_{n})=-\sum_{m=1}^{M}y_{n,m}\ln \hat{y}_{n,m} \label{eq:3},\end{equation} where $y_{n,m}$ and $\hat{y}_{n,m}$ is the ground truth and the prediction result for the $m^\mathsf{th}$ unit of the $n^\mathsf{th}$ training instance, respectively and $M$ is the number of output units.
%However, this is equivalent to maximizing the square of the length of the vector formed by the hashing layer activations, that is $\sum_{n=1}^{N}||\textbf{o}^{H}_{n}-\textbf{0}||^{2}=\sum_{n=1}^{N}||\textbf{o}^{H}_{n}||^{2}$.
In addition to using classification loss and the continuation method described in Sec. \ref{subsec:mdh}, we add constraint of quantization loss and entropy maximization to generate efficient binary codes at the output of the MDH module. The output of the hashing layer is a $J$-dimensional vector denoted by $\textbf{o}^{H}_{n}$, corresponding to the $n$-th input image. The $i$-th element of this vector is denoted by $o^{H}_{n,i} (i=1,2,3, \cdots,J)$. The value of $o^{H}_{n,i}$ is in the range of $[-1,1]$ because it has been activated by the $\mbox{tanh}$ activation. To make the codes closer to either -1 or 1, we add a constraint of quantization loss between the hashing layer activations and 0, which is given by $\sum_{n=1}^{N}||\textbf{o}^{H}_{n}-\textbf{0}||^{2}$, where $N$ is the number of training images in a mini-batch and \textbf{0} is the $J$-dimensional vector with all elements equal to 0.  Let $E_{2}(\textbf{w})$ denote this constraint to boost the activations of the units in the hashing layer to be closer to -1 or 1:
\vspace{-0.15cm}
\begin{equation}E_{2}(\textbf{w})=-\frac{1}{J}\sum_{n=1}^{N}||\textbf{o}^{H}_{n}-\textbf{0}||^{2}=-\frac{1}{J}\sum_{n=1}^{N}||\textbf{o}^{H}_{n}||^{2}  \label{eq:4}.\end{equation}  

In addition to forcing the codes to become binarized, we also require that the binary codes have equal number of -1's and 1's, which maximizes the entropy of the discrete distribution and results in hash codes with better discrimination. Let $E_{3}(\textbf{w})$ denote this constraint that forces the output of each node to have a $50\%$ chance of being -1 or 1:
\vspace{-0.15cm}
\begin{equation}E_{3}(\textbf{w})= \sum_{n=1}^{N}(\text{mean}(\textbf{o}^{H}_{n}))^{2}\label{eq:5}.\end{equation}

Therefore, the overall objective function that is required to be minimized to generate discriminative efficient binary codes is given as:
\begin{equation}\alpha E_{1}(\textbf{w})+ \beta E_{2}(\textbf{w}) + \gamma E_{3}(\textbf{w}) \label{eq:6},\end{equation} where $\alpha$, $\beta$, and $\gamma$ are the tuning parameters of each term.

\subsection{Step 2: Generating the Ground Truth for Training the Neural Network Decoder}

 In \cite{2016_nachmani_NND}, the goal is to optimize the NND for a gaussian noise channel and the database used reflects various channel output realizations when the zero codeword has been transmitted. However, for our proposed system, we want to optimize the NND to be used with biometric data, where the channel noise for our database is characterized by the image distortions (e.g., pose variations, illumination variations and noise due to biometric capturing device) in different biometric images of the same subject. We can create the input dataset for training the NND by using the different facial and iris images for the same subject. However, we do not have the labels or the ground truth codewords that these input images need to be mapped to. An external conventional ECC decoder can be used to generate the ground truth codewords for this input dataset.
 %To generate the ground truth codewords, we use an external conventional ECC decoder
 
 %For this reason, we need to use an external ECC decoder to be used for decoding and generating the codewords for the input biometric images.  
 
 %This step can be avoided and the NND can directly be trained using the method as described in the original paper \cite{2016_nachmani_NND}. In the original paper the database required to train the NND is constructed by considering noisy versions of a zeros codeword, and it reflects various channel output realizations when the zero codeword has been transmitted. The goal in the original paper is to train the parameters of the NND to achieve $N$ dimensional output codeword, which is as close as possible to the zero codeword. 

After training the MDH network in Step 1, facial and iris image pairs for a different set of subjects disjoint from the training dataset are used for generating the ground truth for training the NND. We use the MDH network to extract the joint binary features for this disjoint dataset. These extracted features are used as input to a conventional ECC decoder for soft-decision decoding. Hard-limiting the output of the ECC decoder generates ground truth codewords that are used as labels for optimizing the NND in Step 3. 

For this step, we have used the facial and iris image pairs of 294 common subjects from the 2013 year of the WVU multimodal 2012-2013 dataset. This implies that we have extracted the features for all the face and iris image pairs of the 294 subjects using the MDH network and decoded it using soft-decision decoding with an ECC decoder to generate the ground truth to be used in Step 3. While usually all the joint feature vectors of a given subject are mapped by the decoder to the same codeword, it is possible that the vectors could be mapped to different codewords.  This is especially the case when there are substantive differences in the latent subspaces of the different face/image pairs for the subject or when the ECC code is not sufficiently strong.  In the case that the feature vectors get mapped to a plurality of codewords, the most common of these codewords is used as the ground truth for that subject.

%There could be a possibility that all the joint feature vectors of a given subject may not decode to the same codeword. In this case, we choose the codeword that the maximum number of feature vectors have decoded to, for that subject. 

%For example, let us assume for a given subject $i$, we use 5 image pairs, which gives us 5 joint feature vectors extracted using the MDH network. These feature vectors are passed through the conventional decoder to generate the 5 codewords. Let us assume the 5 codewords to be $x,y,x,x,y$. In this case we will chose $x$ as the ground truth label for the subject $i$. In case if the 5 decoded codewords are$x,y,x,y,z$, then we can chose $x$ or $y$, or we can look at average mean distance of $x$ and $y$ from the other codewords for that subject.          

\subsection{Step 3: Joint Optimization of the Multimodal Deep Hashing Neural Decoder}\label{subsec:joint_optimization}

In Step 3 of the training, first the NND is trained using the procedure and database from the original paper \cite{2016_nachmani_NND}. In the next step, the NND is fine-tuned for our multibiometric data. For fine-tuning the NND, we use the same database used for the ECC decoder in Step 2, where the input to NND is given by the feature vectors generated by the MDH network and the labels are provided by the decoded codewords generated by the conventional ECC decoder in Step 2. Using this database, the NND is fine-tuned for our joint biometric data. We have used sigmoid activation for the last layer of NND. The sigmoid is added so that the final network output is in the range $[0,1]$. This makes it possible to train and fine-tune the NND using cross-entropy loss function:
\begin{equation}
    L(o,y)=-\frac{1}{N}\sum_{i=1}^{N} y_{i}\log(o_i)+(1-y_i)\log(1-o_i), \label{eq:7}
\end{equation} where $o_i$,$y_i$ are  the  deep  neural  network  output  and  the actual $i$th  component  of  the  ground truth codeword (label), respectively. 

After fine-tuning the NND, we integrate MDH and NND by discarding the softmax layer in the MDH network and connecting the output of the hashing layer from MDH as input to NND to create and end-to-end MDHND network. This MDHND is optimized end-to-end using the same dataset as used for fine-tuning NND and also the same cross-entropy loss function given in (\ref{eq:7}).   
For fine-tuning the NND and end-to-end optimization of the MDHND, we use the same 294 subjects from the 2013 year of the WVU multimodal dataset used in Step 2. The total number of facial and iris images corresponding to the 294 subjects is equal to 15,500 and 33, 200, respectively. We use the Adam optimizer \cite{kingma2014adam} with the default hyper-parameter values ($\epsilon = 10^{-3}$, $\beta_1 = 0.9$, $\beta_2 = 0.999$) to train all the parameters. The batch size in all the experiments is fixed to 32. Our MDHND is implemented in TensorFlow with python API and all the experiments are conducted on two GeForce GTX TITAN X 12GB GPUs. 

\section{Performance Evaluation}

\subsection{Details of the Code and Decoder}
The intermediate binary code generated from the MDH module is considered to be the noisy codeword of some error correcting code that  we  can  select and this noisy codeword can be decoded using the NND. We have used BCH code for our experiments. The size of the intermediate binary codes, which is the output of the MDH network depends upon the size of the error correcting code being used for the NND.  We have experimented with a few different sizes of the BCH code for NND including BCH(63,45), BCH(127,85), BCH(255,187), and BCH(511,376). We have tried to keep the code rate to be around $0.7$. For the external ECC decoder, we have used the BCH decoder from the communication toolbox in MATLAB\textsuperscript{\textregistered}. 

\subsection{Evaluation Protocol for Authentication}

For evaluation of authentication performance, we use a disjoint dataset of 70 subjects, which have never been seen during training with 20 facial and 20 iris images per subject leading to a total of 1400 face and iris image pairs with no repetitions. For authentication, all the 1400 pairs are forward passed through our proposed MDHND system and genuine and impostor scores are calculated using Hamming distance as our score measure. Based on the number of subjects ($N=70$) and the number of image pairs ($t=20$) per subject we obtain $Nt(t-1)/2 = 13,300$ genuine scores and $(N(N-1)t^2)/2=966,000$ imposter scores. We have used the ROC curve and the EER as our performance metric for authentication. The ROC curve is plotted between the genuine accept rate (GAR) and the false acceptance rate (FAR) at all possible values of score thresholds. EER indicates a value that the proportion of false acceptances is equal to the proportion of false rejections.

\subsection{Evaluation Protocol for Identification}

 We have also evaluated our system for identification application. The proposed system can also be used for identification as long as the database of individuals is not too large because if we make the ECC too strong, it may lead to ambiguity, where multiple users are resolved to the same codeword and the system may not be too discriminative and may not provide good performance. For identification performance, MDHND is trained on 294 mutual subjects from year 2013, and is tested on the same subjects from year 2012 of the WVU multimodal dataset. The total number of image pairs in the test set for identification is equal to 2940 corresponding to 294 subjects. For identification, the score can be calculated against each of the 294 stored codes and the class yielding the best score can be identified as the subject class. We have used the classification accuracy as our performance metric for identification. The classification accuracy is defined as the ratio of the correct predictions made to the the total number of predictions for a given testing set.    

\subsection{Baselines}
We have compared our algorithms with some of the state-of-the-art score, decision, and fusion-level algorithms. CNN-Sum is a score-level fusion algorithms, which use the probability outputs for the test sample of each modality, added together to give the final score vector. CNN-Major is a decision-level fusion algorithm, which chooses the maximum number of modalities taken to be from the correct class. The feature level fusion techniques include serial feature fusion \cite{chengjun_2001_serial}, parallel feature fusion \cite{Yang_2003_FeatureFP}, CCA-based feature fusion \cite{Sun_2005_new_method}, and discriminant correlation analysis (DCA/MDCA)\cite{Haghiaghat_2016_dca} methods. For the serial, parallel and CCA fusion techniques, we have used principal component analysis (PCA) \cite{mackiewicz_1993_principal} and linear discriminant analysis (LDA) for dimensionality reduction followed by K-nearest neighbor (KNN) classifier \cite{Cover_1967_knn}. To compare the results for the proposed system, with the state-of-the-art algorithms, we extract Gabor features in five scales and eight orientations for face and iris modalities. For each face, and iris image 31, 360, and 36, 630 features are extracted respectively. These hand-crafted Gabor features are used only for serial, parallel, CCA and DCA fusion techniques. For all the other algorithms being compared we have used CNN features. 

We have also compared our multibiometric system with single modality system denoted as Iris-CNN and Face-CNN. As stated previously, we have implemented two different fusion architectures FCA and BLA in our MDH network. MDH-FCA/BLA denotes a stand-alone MDH network with feature level fusion using FCA/BLA and with no ECC decoder of any kind being used. For such cases in comparison, when no error correcting code is used, the length of the output feature vector is equal to the codeword length n of the BCH(n,k) code. MDH-FCA+Ext.Decoder/MDH-BLA+Ext.Decoder denotes feature level fusion using FCA/BLA in the MDH network integrated with an external conventional decoder. MDH-FCA+NND/MDH-FCA+NND denotes feature level fusion using FCA/BLA in the MDH network integrated with a non optimized NND (NND is just trained as in \cite{2016_nachmani_NND} with AWGN channel but not fine-tuned with biometric data) and also with no joint optimization. MDHND-FCA/MDHND-BLA denotes our proposed overall system with feature level fusion using FCA/BLA in the MDH network integrated and jointly optimized with a NND for biometric data. 

%We have also compared the overall system of MDHND with only MDH network and also MDH+NND. In MDH+NND, the NND is just trained as in the original paper without being optimized for biometric data, and also integrated into the MDH with no joint optimization. FCA and BLA indicate the two architectures that have been used for feature-level fusion of individual modalities in the MDH network. Additionally, we have also compared the performance with modality specific face and iris CNNs.  

\subsection{Authentication Results}

\begin{table}[b]
\centering
\small
\scalebox{0.70}{\begin{tabular}{|c|c|c|c|c|}
 \hline
\multicolumn{1}{|c}{\multirow{1}{*}{$Algorithm$}} &\multicolumn{1}{|c}{\multirow{1}{*}{BCH(63,45)}}
&\multicolumn{1}{|c}{\multirow{1}{*}{BCH(127,85)}} &\multicolumn{1}{|c}{\multirow{1}{*}{BCH(255,187)}}
&\multicolumn{1}{|c|}{BCH(511,376)}\\ \hline
\multicolumn{1}{|c}{\multirow{1}{*}{CNN-Sum}} &\multicolumn{1}{|c}{\multirow{1}{*}{1.71\%}}
&\multicolumn{1}{|c}{\multirow{1}{*}{1.64\%}} &\multicolumn{1}{|c}{\multirow{1}{*}{1.32\%}}
&\multicolumn{1}{|c|}{1.29\%}\\ 
\multicolumn{1}{|c}{\multirow{1}{*}{CNN-Major}} &\multicolumn{1}{|c}{\multirow{1}{*}{2.11\%}}
&\multicolumn{1}{|c}{\multirow{1}{*}{2.19\%}} &\multicolumn{1}{|c}{\multirow{1}{*}{1.61\%}}
&\multicolumn{1}{|c|}{1.53\%}\\
\multicolumn{1}{|c}{\multirow{1}{*}{Iris-CNN}} &\multicolumn{1}{|c}{\multirow{1}{*}{2.74\%}}
&\multicolumn{1}{|c}{\multirow{1}{*}{2.34\%}} &\multicolumn{1}{|c}{\multirow{1}{*}{2.10\%}}
&\multicolumn{1}{|c|}{2.04\%}\\
\multicolumn{1}{|c}{\multirow{1}{*}{Face-CNN}} &\multicolumn{1}{|c}{\multirow{1}{*}{1.91\%}}
&\multicolumn{1}{|c}{\multirow{1}{*}{1.72\%}} &\multicolumn{1}{|c}{\multirow{1}{*}{1.63\%}}
&\multicolumn{1}{|c|}{1.58\%}\\
\multicolumn{1}{|c}{\multirow{1}{*}{MDH-FCA}} &\multicolumn{1}{|c}{\multirow{1}{*}{1.84\%}}
&\multicolumn{1}{|c}{\multirow{1}{*}{1.67\%}} &\multicolumn{1}{|c}{\multirow{1}{*}{1.48\%}}
&\multicolumn{1}{|c|}{1.43\%}\\ 
\multicolumn{1}{|c}{\multirow{1}{*}{MDH-BLA}} &\multicolumn{1}{|c}{\multirow{1}{*}{1.76\%}}
&\multicolumn{1}{|c}{\multirow{1}{*}{1.66\%}}
&\multicolumn{1}{|c}{\multirow{1}{*}{1.45\%}}
&\multicolumn{1}{|c|}{1.42\%}\\
\multicolumn{1}{|c}{\multirow{1}{*}{MDH-FCA+Ext.Decoder}} &\multicolumn{1}{|c}{\multirow{1}{*}{1.73\%}}
&\multicolumn{1}{|c}{\multirow{1}{*}{1.64\%}} &\multicolumn{1}{|c}{\multirow{1}{*}{1.41\%}}
&\multicolumn{1}{|c|}{1.39\%}\\ 
\multicolumn{1}{|c}{\multirow{1}{*}{MDH-BLA+Ext.Decoder}} &\multicolumn{1}{|c}{\multirow{1}{*}{1.62\%}}
&\multicolumn{1}{|c}{\multirow{1}{*}{1.58\%}}
&\multicolumn{1}{|c}{\multirow{1}{*}{1.32\%}}
&\multicolumn{1}{|c|}{1.28\%}\\ 
\multicolumn{1}{|c}{\multirow{1}{*}{MDH-FCA+NND}} &\multicolumn{1}{|c}{\multirow{1}{*}{1.39\%}}
&\multicolumn{1}{|c}{\multirow{1}{*}{1.34\%}} &\multicolumn{1}{|c}{\multirow{1}{*}{1.30\%}}
&\multicolumn{1}{|c|}{1.23\%}\\ 
\multicolumn{1}{|c}{\multirow{1}{*}{MDH-BLA+NND}} &\multicolumn{1}{|c}{\multirow{1}{*}{1.30\%}}
&\multicolumn{1}{|c}{\multirow{1}{*}{1.26\%}} &\multicolumn{1}{|c}{\multirow{1}{*}{1.24\%}}
&\multicolumn{1}{|c|}{1.19\%}\\ 
\multicolumn{1}{|c}{\multirow{1}{*}{\textbf{MDHND-FCA}}} &\multicolumn{1}{|c}{\multirow{1}{*}{\textbf{1.05\%}}}
&\multicolumn{1}{|c}{\multirow{1}{*}{\textbf{0.99\%}}} &\multicolumn{1}{|c}{\multirow{1}{*}{\textbf{0.90\%}}}
&\multicolumn{1}{|c|}{\textbf{0.86\%}}\\ 
\multicolumn{1}{|c}{\multirow{1}{*}{\textbf{MDHND-BLA}}} &\multicolumn{1}{|c}{\multirow{1}{*}{\textbf{1.01\%}}}
&\multicolumn{1}{|c}{\multirow{1}{*}{\textbf{0.94\%}}} &\multicolumn{1}{|c}{\multirow{1}{*}{\textbf{0.84\%}}}
&\multicolumn{1}{|c|}{\textbf{0.79\%}}\\ \hline
\end{tabular}}
\vspace{0.15cm}
\caption{EER for different algorithms using different BCH codes.}
\label{table:eer_comp}
\vspace{-0.7cm}
\end{table}

Table \ref{table:eer_comp} shows a comparison of authentication performance for our proposed system with other methods in terms of EER. Our proposed systems (MDHND-FCA/BLA) are shown in bold in the last 2 rows of the table. As it can be seen that our proposed MDHND framework outperforms single modality by at least $1.5\%$. It can be observed that there is an improvement in  performance  by  using  additional  biometric  features  and the  multimodality  system performs  better  than  the unimodal systems. We can also observe that there is an improvement in performance when ECC decoding is used, and that this improvement improves when the NND decoder is used. For example, for BCH(255,187) and the BLA architecture, the EER decreases from $1.45\%$ without ECC decoding to $1.32\%$ when a conventional decoder is used.  The EER further decreases to $1.24\%$ when an unoptimized NND is used (i.e., trained on AWGN) and to $0.84\%$ when an optimized NND is used (i.e., trained on biometric data) for a net improvement of about $0.5\%$ relative to the conventional decoder. 

%It can also be observed that an increase in the size of the codeword  also helps to improves the authentication performance. This can be due to improvement in error correcting capability of the code 

\begin{figure}[t]

\centering
\includegraphics[width=7.6cm]{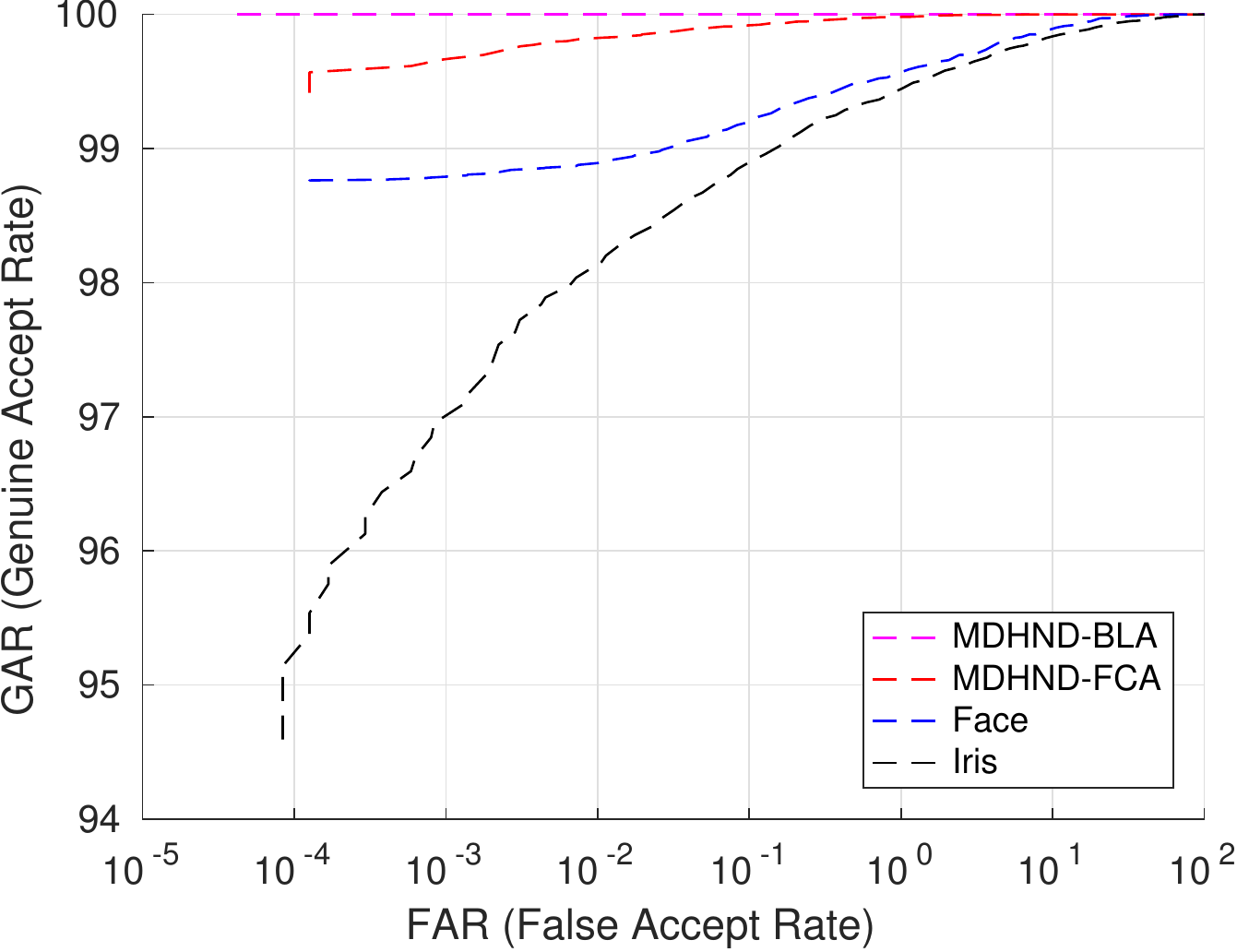}
\vspace{-0.2cm}
\caption{ROC curves for face, iris, MDHND-FCA, and MDHND-BLA modalities using BCH(511,376) }\label{fig:ROC_fca_bla_bch_255} 
\vspace{-0.1cm}
\end{figure}

\begin{figure}[t]

\centering
\includegraphics[width=7.6cm]{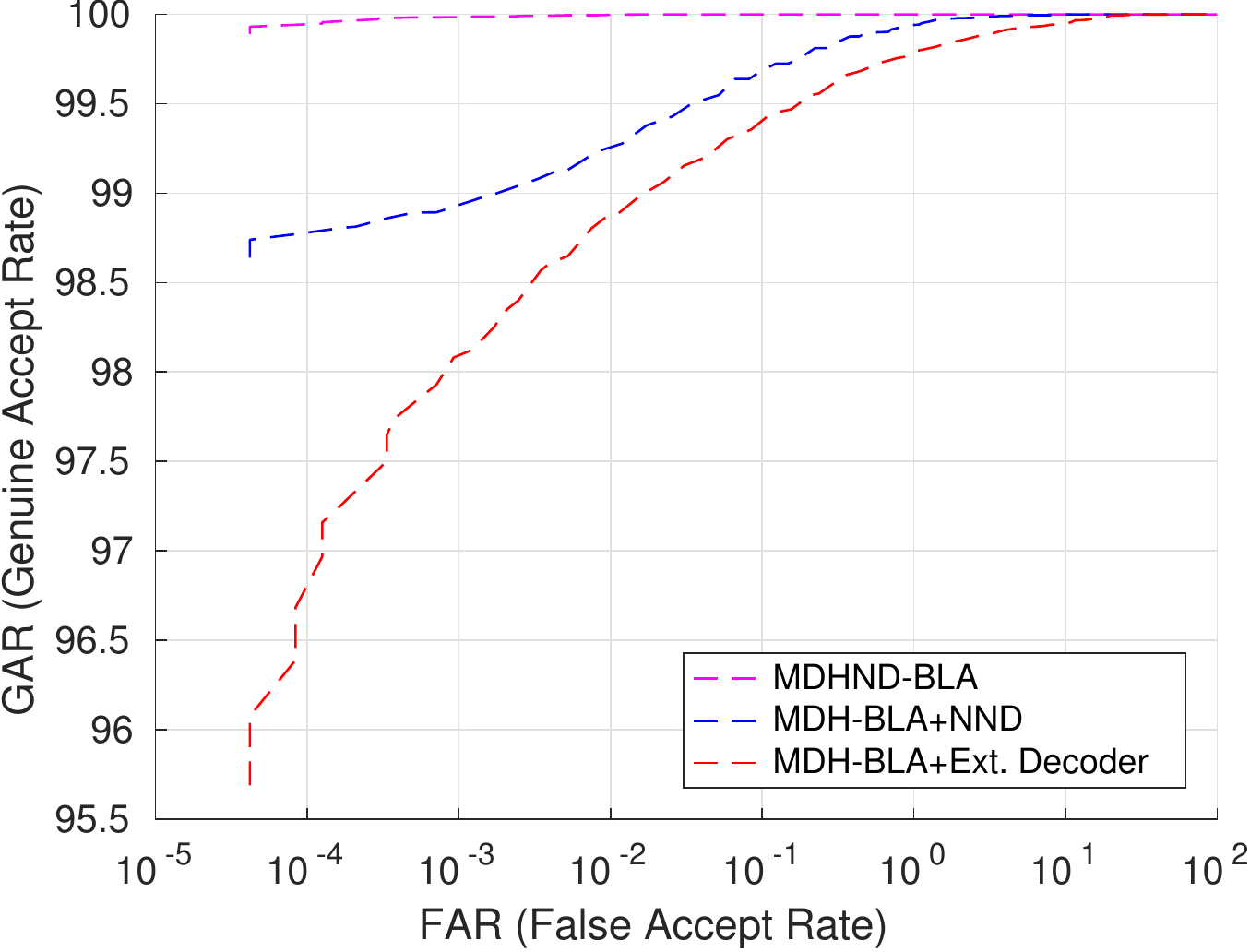}
\vspace{-0.2cm}
\caption{ROC curves for face, iris, MDHND-FCA, and MDHND-BLA modalities using BCH(255,187) and BLA architecture }\label{fig:ROC_nnd} 
\vspace{-0.1cm}
\end{figure}

Fig. \ref{fig:ROC_fca_bla_bch_255} shows the comparison of ROC curves for unimodal and a multimodal system. It is evident that the multimodal feature-level fusion using BLA integrated with NND and joint optimization significantly improves unimodal representation accuracy by using a bilinear formulation for exploiting the captured multiplicative interactions of the low-dimensional modality-dedicated feature representations. Fig. \ref{fig:ROC_nnd} shows the comparison of ROC curves showing the improvement in performance by using an optimized NND (MDHND-BLA) relative to a non-optimized NND (MDH-BLA+NND) and an external decoder (MDH-BLA+Ext. Decoder). It can clearly be seen that with a FAR of $0.01\%$ we get an improvement in GAR of about $1.3\%$ by using an optimized NND relative to a conventional external decoder. 

We have also evaluated the time/delay required to perform a single authentication. We have measured the time for 1000 authentications using our trained MDHND-BLA with the same hardware as described at the end of section \ref{subsec:joint_optimization} . Based on this evaluation, the total time required for 1000 authentication is equal to 9.625 secs, which equals to an average of 9.625ms per authentication.  

\balance
\subsection{Identification Results}

\begin{table}
\centering
\small
\scalebox{0.70}{\begin{tabular}{|c|c|c|c|c|}
 \hline
\multicolumn{1}{|c}{\multirow{1}{*}{$Algorithm$}} &\multicolumn{1}{|c}{\multirow{1}{*}{BCH(63,45)}}
&\multicolumn{1}{|c}{\multirow{1}{*}{BCH(127,85)}} &\multicolumn{1}{|c}{\multirow{1}{*}{BCH(255,187)}}
&\multicolumn{1}{|c|}{BCH(511,376)}\\ \hline 
\multicolumn{1}{|c}{\multirow{1}{*}{Serial + PCA + KNN}} &\multicolumn{1}{|c}{\multirow{1}{*}{68.92\%}}
&\multicolumn{1}{|c}{\multirow{1}{*}{70.47\%}} &\multicolumn{1}{|c}{\multirow{1}{*}{70.83\%}}
&\multicolumn{1}{|c|}{71.12\%}\\ 
\multicolumn{1}{|c}{\multirow{1}{*}{Serial + LDA + KNN}} &\multicolumn{1}{|c}{\multirow{1}{*}{76.11\%}}
&\multicolumn{1}{|c}{\multirow{1}{*}{78.00\%}} &\multicolumn{1}{|c}{\multirow{1}{*}{80.15\%}}
&\multicolumn{1}{|c|}{80.52\%}\\ 
\multicolumn{1}{|c}{\multirow{1}{*}{Parallel + PCA + KNN}} &\multicolumn{1}{|c}{\multirow{1}{*}{71.2\%}}
&\multicolumn{1}{|c}{\multirow{1}{*}{73.21\%}} &\multicolumn{1}{|c}{\multirow{1}{*}{74.1\%}}
&\multicolumn{1}{|c|}{74.69\%}\\ 
\multicolumn{1}{|c}{\multirow{1}{*}{Parallel + LDA + KNN}} &\multicolumn{1}{|c}{\multirow{1}{*}{77.4\%}}
&\multicolumn{1}{|c}{\multirow{1}{*}{80.19\%}} &\multicolumn{1}{|c}{\multirow{1}{*}{82.11\%}}
&\multicolumn{1}{|c|}{82.53\%}\\ 
\multicolumn{1}{|c}{\multirow{1}{*}{CCA + PCA + KNN}} &\multicolumn{1}{|c}{\multirow{1}{*}{84.34\%}}
&\multicolumn{1}{|c}{\multirow{1}{*}{85.12\%}} &\multicolumn{1}{|c}{\multirow{1}{*}{87.45\%}}
&\multicolumn{1}{|c|}{87.21\%}\\ 
\multicolumn{1}{|c}{\multirow{1}{*}{CCA + LDA + KNN}} &\multicolumn{1}{|c}{\multirow{1}{*}{87.43\%}}
&\multicolumn{1}{|c}{\multirow{1}{*}{88.65\%}} &\multicolumn{1}{|c}{\multirow{1}{*}{88.98\%}}
&\multicolumn{1}{|c|}{89.12\%}\\ 
\multicolumn{1}{|c}{\multirow{1}{*}{DCA/MDCA + KNN}} &\multicolumn{1}{|c}{\multirow{1}{*}{79.08\%}}
&\multicolumn{1}{|c}{\multirow{1}{*}{81.67\%}} &\multicolumn{1}{|c}{\multirow{1}{*}{82.09\%}}
&\multicolumn{1}{|c|}{83.02\%}\\ \hline\hline
\multicolumn{1}{|c}{\multirow{1}{*}{CNN-Sum}} &\multicolumn{1}{|c}{\multirow{1}{*}{95.13\%}}
&\multicolumn{1}{|c}{\multirow{1}{*}{95.21\%}} &\multicolumn{1}{|c}{\multirow{1}{*}{95.54\%}}
&\multicolumn{1}{|c|}{96.11\%}\\ 
\multicolumn{1}{|c}{\multirow{1}{*}{CNN-Major}} &\multicolumn{1}{|c}{\multirow{1}{*}{93.34\%}}
&\multicolumn{1}{|c}{\multirow{1}{*}{94.16\%}} &\multicolumn{1}{|c}{\multirow{1}{*}{95.23\%}}
&\multicolumn{1}{|c|}{95.71\%}\\
\multicolumn{1}{|c}{\multirow{1}{*}{Iris-CNN}} &\multicolumn{1}{|c}{\multirow{1}{*}{92.03\%}}
&\multicolumn{1}{|c}{\multirow{1}{*}{92.39\%}} &\multicolumn{1}{|c}{\multirow{1}{*}{93.22\%}}
&\multicolumn{1}{|c|}{94.91\%}\\
\multicolumn{1}{|c}{\multirow{1}{*}{Face-CNN}} &\multicolumn{1}{|c}{\multirow{1}{*}{94.19\%}}
&\multicolumn{1}{|c}{\multirow{1}{*}{94.97\%}} &\multicolumn{1}{|c}{\multirow{1}{*}{95.06\%}}
&\multicolumn{1}{|c|}{95.65\%}\\
\multicolumn{1}{|c}{\multirow{1}{*}{MDH-FCA}} &\multicolumn{1}{|c}{\multirow{1}{*}{94.76\%}}
&\multicolumn{1}{|c}{\multirow{1}{*}{95.45\%}} &\multicolumn{1}{|c}{\multirow{1}{*}{95.82\%}}
&\multicolumn{1}{|c|}{96.01\%}\\ 
\multicolumn{1}{|c}{\multirow{1}{*}{MDH-BLA}} &\multicolumn{1}{|c}{\multirow{1}{*}{94.88\%}}
&\multicolumn{1}{|c}{\multirow{1}{*}{95.26\%}} &\multicolumn{1}{|c}{\multirow{1}{*}{95.98\%}}
&\multicolumn{1}{|c|}{96.36\%}\\ 
\multicolumn{1}{|c}{\multirow{1}{*}{MDH-FCA+Ext.Decoder}} &\multicolumn{1}{|c}{\multirow{1}{*}{95.16\%}}
&\multicolumn{1}{|c}{\multirow{1}{*}{95.88\%}} &\multicolumn{1}{|c}{\multirow{1}{*}{96.11\%}}
&\multicolumn{1}{|c|}{96.23\%}\\ 
\multicolumn{1}{|c}{\multirow{1}{*}{MDH-BLA+Ext.Decoder}} &\multicolumn{1}{|c}{\multirow{1}{*}{95.22\%}}
&\multicolumn{1}{|c}{\multirow{1}{*}{95.34\%}} &\multicolumn{1}{|c}{\multirow{1}{*}{96.28\%}}
&\multicolumn{1}{|c|}{96.72\%}\\ 
\multicolumn{1}{|c}{\multirow{1}{*}{MDH-FCA+NND}} &\multicolumn{1}{|c}{\multirow{1}{*}{96.32\%}}
&\multicolumn{1}{|c}{\multirow{1}{*}{97.22\%}} &\multicolumn{1}{|c}{\multirow{1}{*}{97.28\%}}
&\multicolumn{1}{|c|}{97.83\%}\\
\multicolumn{1}{|c}{\multirow{1}{*}{MDH-BLA+NND}} &\multicolumn{1}{|c}{\multirow{1}{*}{96.98\%}}
&\multicolumn{1}{|c}{\multirow{1}{*}{97.11\%}} &\multicolumn{1}{|c}{\multirow{1}{*}{97.29\%}}
&\multicolumn{1}{|c|}{98.12\%}\\ 
\multicolumn{1}{|c}{\multirow{1}{*}{\textbf{MDHND-FCA}}} &\multicolumn{1}{|c}{\multirow{1}{*}{\textbf{98.02\%}}}
&\multicolumn{1}{|c}{\multirow{1}{*}{\textbf{98.13\%}}} &\multicolumn{1}{|c}{\multirow{1}{*}{\textbf{99.10\%}}}
&\multicolumn{1}{|c|}{\textbf{99.11\%}}\\ 
\multicolumn{1}{|c}{\multirow{1}{*}{\textbf{MDHND-BLA}}} &\multicolumn{1}{|c}{\multirow{1}{*}{\textbf{98.16\%}}}
&\multicolumn{1}{|c}{\multirow{1}{*}{\textbf{98.94\%}}} &\multicolumn{1}{|c}{\multirow{1}{*}{\textbf{99.13\%}}}
&\multicolumn{1}{|c|}{\textbf{99.23\%}}\\ \hline
\end{tabular}}\vspace{0.05cm}
\caption{Classification accuracy using different BCH codes.}
\label{table:acc_comp}
\vspace{-0.7cm}
\end{table}

Table \ref{table:acc_comp} shows a comparison of our proposed systems with different state of the art fusion algorithms for identification task. It can clearly be seen that the performance of our proposed systems (MDHND-FCA and MDHND-BLA shown in the last two rows) in  terms  of  classification accuracy is significantly  better  than  previous fusion approaches. It can also be observed that using optimized NND (MDHND-BLA) definitely helps to improve the identification performance and we get an improvement of about $3\%$ when compared to a conventional external ECC decoder (MDH-BLA+Ext.Decoder) or an improvement of about $3.5\%$ with no decoder (MDH-BLA) being used at all.

\section{Conclusion and Future Work}

 We  have  presented  a feature-level  fusion  and  binarization framework  using  deep  hashing, and integrated a neural network decoder into the framework. The result is a design for a multimodal  biometric system that leverages each  user's  multiple  biometrics for authentication. In this framework, we leveraged a neural network  based  decoder  to  refine the codes  generated  by the deep hashing network to improve the authentication performance.  We have also implemented multiple architectures for combining the biometrics at feature-level. The experimental results show that the bilinear architecture is better than linear concatenation of features. Additionally, the experimental results show that the optimized neural network decoder decreases the EER of the multimodal biometric system by $0.7\%$, relative to not using any decoder at all. Also, optimized neural network decoder significantly improves the authentication performance (GAR) of the multimodal biometric system by about $1.3\%$ at an FAR of $0.01\%$, when compared to using an external conventional ECC decoder. The current work deals with fusion of two modalities and we plan to extend our model and use more than two modalities. Also we intend to use more recent codes such as Turbo and LDPC codes as our error-correcting code. 
 
 %Moreover, the results indicate that the proposed framework outperforms most of the other feature-level fusion approaches.

{\small
\bibliographystyle{IEEEtran}
\bibliography{icc}
}

\end{document}